\documentclass[letterpaper]{article} 
\usepackage{aaai25}  
\usepackage{times}  
\usepackage{helvet}  
\usepackage{courier}  
\usepackage[hyphens]{url}  
\usepackage{graphicx} 
\usepackage{natbib}  
\usepackage{caption} 
\usepackage{algorithm}
\usepackage{algorithmic}
\usepackage{amsmath,amssymb,amsfonts}
\usepackage{newfloat}
\usepackage{listings}
\DeclareCaptionStyle{ruled}{labelfont=normalfont,labelsep=colon,strut=off} 
\floatstyle{ruled}
\newfloat{listing}{tb}{lst}{}
\floatname{listing}{Listing}
\title{SpikeGS: Reconstruct 3D Scene Captured by a Fast-Moving
Bio-Inspired Camera}

\author{Yijia Guo\textsuperscript{\rm 1}\equalcontrib, 
Liwen Hu\textsuperscript{\rm 1}\equalcontrib,
 Yuanxi Bai\textsuperscript{\rm 2},
Jiawei Yao\textsuperscript{\rm 3},
Lei Ma\textsuperscript{\rm 1,2}\thanks{Corresponding author.
},
Tiejun Huang\textsuperscript{\rm 1}\\
}
\affiliations{\textsuperscript{\rm 1}National Key Laboratory for Multimedia Information Processing,  School of Computer Science, Peking University\\
\textsuperscript{\rm 2}National Biomedical Imaging Center, Peking University\\
\textsuperscript{\rm 3}University of Washington\\
\{2301112015, 230121059\}@Stu.pku.edu.cn, \{liwenhu, leima, tjhuang\}@pku.edu.cn, 
jwyao@uw.edu
}

\usepackage{bibentry}

\begin{document}

\maketitle


%
\begin{abstract}
3D Gaussian Splatting (3DGS) has been proven to exhibit exceptional performance in reconstructing 3D scenes. However, the effectiveness of 3DGS heavily relies on sharp images, and fulfilling this requirement presents challenges in real-world scenarios particularly when utilizing fast-moving cameras. This limitation severely constrains the practical application of 3DGS and may compromise the feasibility of real-time reconstruction.  
To mitigate these challenges, we proposed Spike Gaussian Splatting (SpikeGS), the first framework that integrates the Bayer-pattern spike streams into the 3DGS
pipeline to reconstruct 3D scenes captured by a fast-moving high temporal resolution color spike camera \textbf{in one second}. 
With accumulation rasterization, interval supervision, and a special designed pipeline, SpikeGS realizes continuous spatiotemporal perception while extracts detailed structure and texture from Bayer-pattern spike stream which is unstable and lacks details. 
Extensive experiments on both synthetic and real-world datasets
demonstrate the superiority of SpikeGS compared with existing spike-based and deblurring 3D scene reconstruction methods. 
\begin{links}
\link{Code}{https://spikegs.github.io.}
\end{links}
\end{abstract}
%
%
    
\vspace{-0.3cm}
\section{Introduction}
\label{sec:intro3}
Speed remains a critical challenge in the area of 3D reconstruction. The achievement of Neural Radiance Field \cite{nerf} has been notably successful. However, the training and rendering processes associated with the neural implicit representation entail a substantial time commitment, resulting in a significant bottleneck. 3DGS \cite{3dgs} has effectively addressed this challenge, remarkably improving reconstruction speeds to minutes and achieving real-time rendering, while InstantSplat \cite{fan2024instantsplat} has further reduced reconstruction time to seconds. However, ultra-fast reconstruction speeds do not inherently guarantee a real-time reconstruction pipeline. Both 3DGS and the majority of methods relying on 3DGS, such as InstantSplat, require sharp image inputs, resulting in the necessity for slow camera movements to prevent motion blur. Capturing even the most basic scenes often needs minutes or more. In the near future, the speed of camera capture, rather than training speed, may become the limiting factor in 3D reconstruction. Presently, the demand for sharp images also limits the application of 3DGS in scenarios when cameras are mounted on high-speed vehicles, e.g., trains and UAVs.
\begin{figure}[htbp]
           \includegraphics[width=1.0\columnwidth]{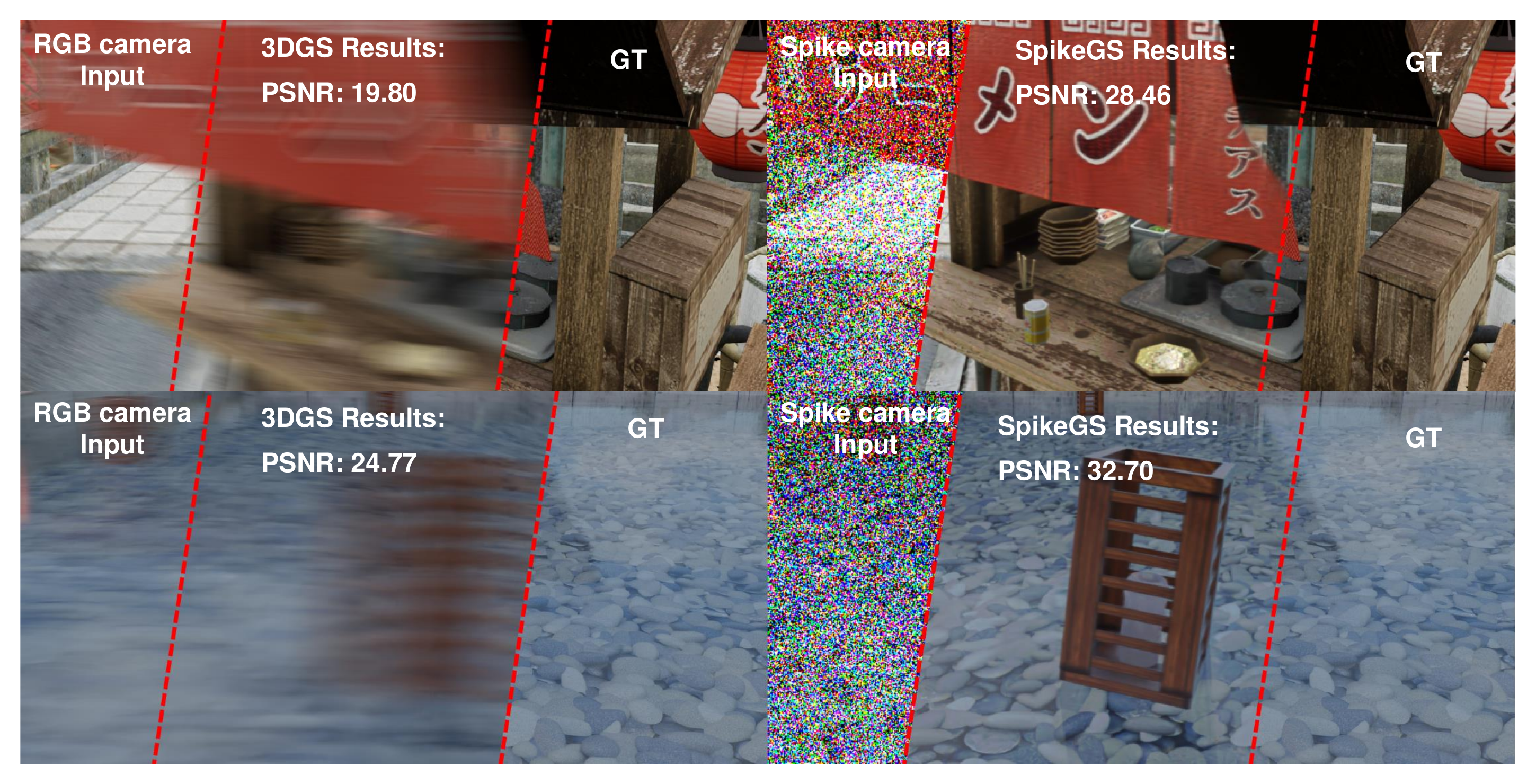}
           \captionof{figure}{Brief introduction of SpikeGS. Given high temporal resolution but texture lacking spike stream captured by a fast-moving spike camera \textbf{in one second}, our method can reconstruct a sharp radiance field in minutes and achieves clear real-time rendering results.}
           \label{fig1}
\end{figure}
Although many deblurring methods \cite{chen2024deblur,zhao2024bad} attempt to address these challenges and free the constraint on camera motion, the limitations of traditional RGB camera sensors greatly hinder their ability to mitigate motion blur effectively. Bio-inspired cameras, characterized by their high temporal resolution, are considered promising solutions to address these issues at their core. Bio-inspired cameras such as event and spike \cite{huang20231000} cameras have been widely applied in 3D reconstruction \cite{rudnev2023eventnerf,xiong2024event3dgs,zhu2024spikenerf} of high-speed scenes with significant success. However, event-based camera approaches may lack adequate visual texture and frequently rely on supplementary traditional cameras for reconstruction \cite{qi2023e2nerf}, thereby constraining their applicability. Spike cameras, which capture rich visual textures, are more suitable for independent 3D reconstruction. However, prior solutions \cite{guo2024spike,zhu2024spikenerf} have encountered difficulties in training and rendering efficiency, which posing an insurmountable challenge for real-time rendering and fast reconstruction. These methods also have not undergone extensive testing in real-world high-speed scenarios and have encountered challenges in reconstructing  Bayer-pattern spike streams as the first-generation spike camera can only reproduce grayscale signals. 
Besides, as depicted in Figure \ref{fig1}, the Bayer-pattern spike stream lacks both geometric consistency and time-domain stability. The rapid reconstruction of precise structures and accurate irradiance from unstable spike streams poses a significant challenge.

In this paper, we combine spike streams with 3DGS to devise a tailored 3D reconstruction pipeline, achieving the reconstruction of scenes captured by spike cameras in one second for the first time. To reconstruct detailed structures and textures from high-temporal resolution but with noisy visual texture spike streams,  we carefully designed accumulation rasterization and interval supervision to enhance continuous spatiotemporal perception. Specifically, accumulation rasterization effectively utilizes long-term temporal information to mitigate temporal aliasing and motion blur, thereby facilitating precise per-pixel detail restoration and achieving sharp reconstruction results. In parallel, interval supervision contributes to the recovery of intricate geometric structures and enables rapid initialization of point clouds and camera poses.
To validate our approach, we have created the first dataset specifically for 3D reconstruction from high-speed motion spike streams, which also serves as the first dataset for 3D reconstruction from Bayer-pattern spike streams. Our experiments demonstrate that our method not only achieves state-of-the-art 3D reconstruction quality for high-speed scenes from spike streams but also exhibits exceptional robustness to speed variations. SpikeGS is also compatible with the first-generation spike camera, as delineated in the appendix. Our main contributions are as follows:
\begin{itemize}
\item[$\bullet$] We propose SpikeGS, \textbf{the first framework} integrating Bayer-pattern spike streams into the 3DGS pipeline for reconstructing 3D scenes using a fast-moving spike camera. Our method is tailored for spike-based supervision, ensuring high robustness across various speeds.
\item[$\bullet$]We contribute SpikeGS dataset, \textbf{the first dataset} for 3D reconstruction from high-speed real-world spike scenes, \textbf{the first dataset} for 3D reconstruction from color spike camera, and \textbf{the first dataset} featuring explicit speed annotations and varying speeds within the same scene. We also provide a standardized paradigm for capturing indoor high-speed spike 3D reconstruction scenes.
\item[$\bullet$] Our experiments demonstrated that our method not only achieves state-of-the-art 3D reconstruction quality for high-speed scenes compared to all current methods but also exhibits exceptional robustness to speed variations. We also validated and proved the feasibility of spike camera reconstruction of open outdoor scenes using synthetic datasets for the first time.
\end{itemize}
\section{Related Work}
\label{sec:related_work}
\subsection{Novel View Synthesis on Traditional Cameras}
Novel View Synthesis (NVS) is the process of generating
new images from viewpoints different from those of the original captures. NeRF-based approaches \cite{nerf,barron2021mipnerf,barron2022mipnerf360,wang2021nerf--,zhang2020nerf++} have become a standard technique in this field and demonstrate that implicit radiance fields can effectively learn scene representations and synthesize high-quality novel views. Further advancements \cite{muller2022instant,chen2022tensorf,fridovich2022plenoxels} have been made to improve the training and rendering of NeRF with advanced implicit scene representations. 3D Gaussian splatting \cite{3dgs} further adopts a discrete 3D Gaussian representation of scenes, significantly accelerating the training and rendering of radiance fields with explicit scenes. It has attracted considerable
research interest in the field of generation \cite{yi2023gaussiandreamer,chen2023gaussianeditor}, relighting \cite{liang2023gsir,guo2024prtgs} and dynamic 3D scene reconstruction \cite{wu20234d,yang2023real}. 

\begin{figure*}[htbp]
\vspace{-0.7cm}
\includegraphics[width=1.0\linewidth]{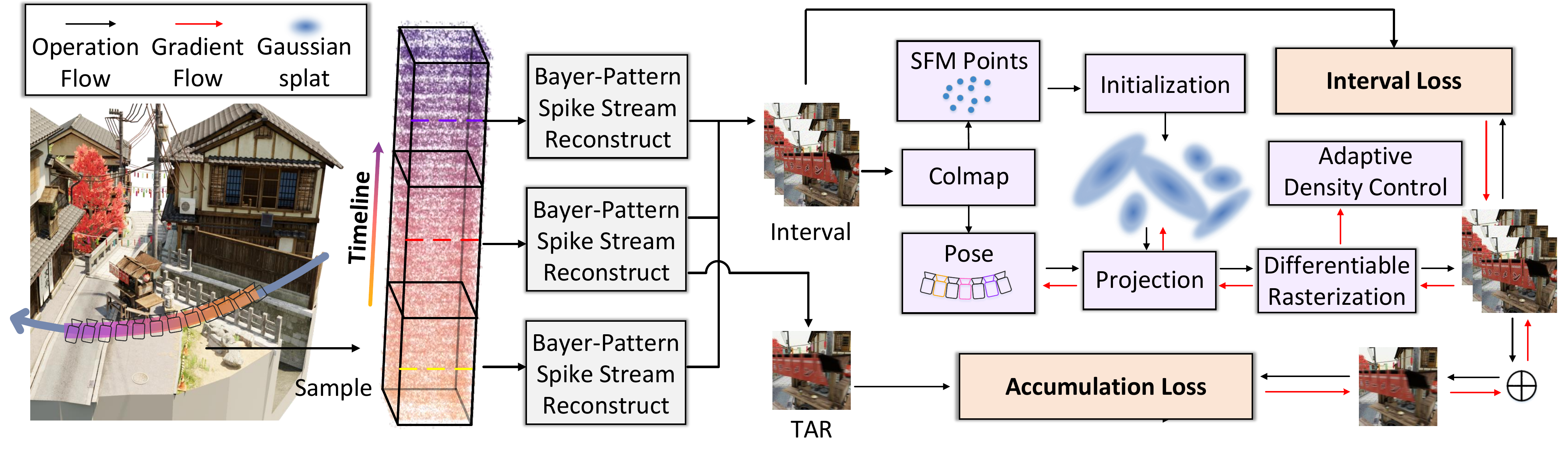}
\centering

\caption{Overview of our SpikeGS. We first reconstruct Bayer-pattern spike streams into spike intervals and spike accumulation (details are shown in Fig.~\ref{tfi_tfp}).
Unlike 3DGS, we adopt spike intervals to initialize SFM points, camera poses, and Gaussian
splats. We then embed the time accumulation process into the rasterizer to calibrate the colorization while maintaining multi-view
consistency. By progressively optimizing the 3DGS parameters using an accumulation loss and an interval loss, our method
facilitates high-quality 3DGS reconstruction.}\label{pipeline}

\end{figure*}

\subsection{Novel View Synthesis on Bio-inspired Cameras}
Bio-inspired sensors have shown their advantages in most computer version problems including novel views synthesizing. EventNeRF \cite{rudnev2023eventnerf} and Ev-nerf \cite{hwang2023ev} synthesize the novel view in scenarios such as high-speed movements that would not be conceivable with a traditional camera by event supervision. Nonetheless, these works assume that event streams are temporally dense and low-noise which is inaccessible in practice. Robust e-NeRF \cite{low2023robust} incorporates a more realistic event generation model to directly and robustly reconstruct NeRFs under various real-world conditions. DE-NeRF \cite{ma2023deformable} and E2NeRF \cite{qi2023e2nerf} extend event nerf to dynamic scenes and severely blurred images as NeRF researchers did. Recently, many researchers \cite{xiong2024event3dgs} introduced 3D Gaussian Splatting framework for reconstructing appearance and geometry
solely from event data, significantly improved the speed and quality of reconstruction. 
 However, due to the absence of texture details in event data, all these approaches yield limited results. Spike cameras \cite{dong2021spike} can capture more texture information. NeRF based on spike cameras \cite{guo2024spike,zhu2024spikenerf} achieves higher 3D scene reconstruction quality than event-based methods. However, the above methods still cannot quickly and efficiently reconstruct real-world 3D scenes while the camera moves at high speeds. They also have problems in dealing with Bayer-pattern spike streams from color spike camera.

\section{Datasets}
\label{sec:related_work}
\begin{figure}[htbp]

\includegraphics[width=\linewidth]{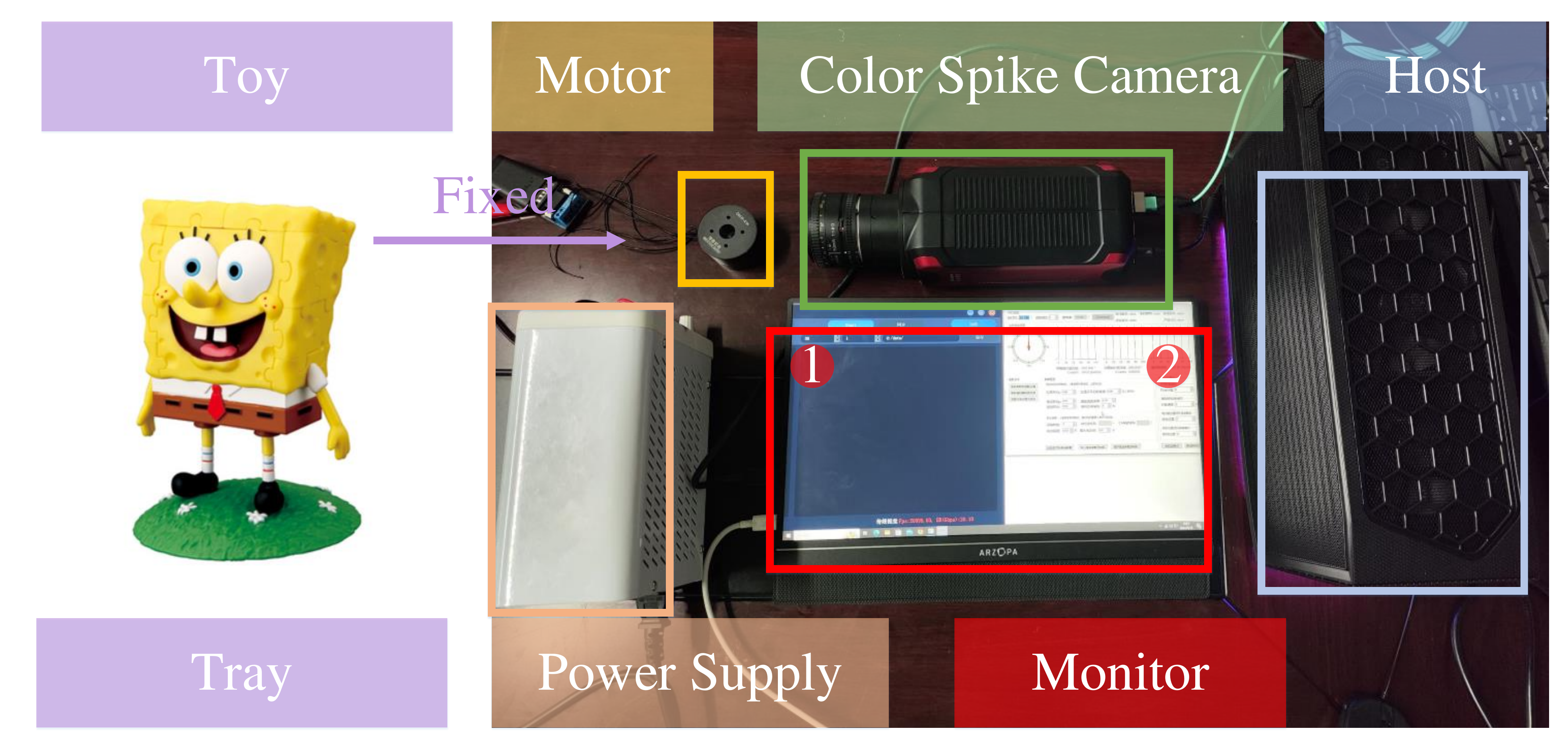}

\centering
\caption{The setup for capturing the real-world dataset, SpikeGS-dataset. First, we fix the toy onto a tray, then the tray is fixed to a motor. Finally, we use a spike camera to capture the high-speed rotating toy. \normalsize{\textcircled{\scriptsize{1}}}\normalsize (\normalsize{\textcircled{\scriptsize{2}}}\normalsize) controls the sampling of the spike camera  (the rotation speed of the motor).}\label{setup}

\end{figure}
\subsection{Synthetic SpikeGS Dataset}
We generate spike streams using six synthetic scenes from NeRF \cite{nerf} and three synthetic scenes from Deblur-nerf \cite{ma2022deblur}. Specifically, we first adjust the camera frame rate in Blender to match the frame rate of the spike camera and render images. Further, we use the state-of-the-art spike camera simulator SCSim \cite{hu2024scsimrealisticspikecameras} to convert the image sequences into Bayer-pattern spike streams. Besides, to investigate the 3D reconstruction performance under different speeds, we vary the motion speed of objects in each scene and generate another set of spike streams with higher speeds. As a result, the high-speed spike streams have a size of $1000 \times 1000 \times 1500$, which indicates that the motion speed is about 2500 rpm. The low-speed spike streams have a size of $1000 \times 1000 \times 2500$, which indicates that the motion speed is about 1500 rpm.

\subsection{Real-world SpikeGS Dataset}

In fact, there is still a significant gap between synthetic and real data for bio-inspired cameras. Therefore, constructing a real 3D high-speed scene dataset is crucial for the evaluation of methods. As shown in Fig.~\ref{setup}, we carefully design an experimental setup to capture spike streams. Based on the setup, we first propose a 3D high-speed scene dataset for color spike cameras. SpikeGS dataset contains a total of 3$\times$7 scenes, where 3 represents 3 different motion speeds (700 rpm, 1000 rpm, 1400 rpm) and 7 represents 7 different objects. For each dynamic scene, we capture a one-second spike stream and its size is $20000 \times 1000 \times 1000$. 


\section{Method}
\label{sec:method}
\subsection{Preliminary:3D Gaussian Splatting}
\label{sec:related_work}
3D Gaussian Splatting constitutes an explicit 3D scene representation utilizing point clouds, with Gaussian splats employed to depict the scene's structure. In this representation, every Gaussian splat $G$ is parameterized by an anisotropic covariance $\Sigma \in \mathbb{R}^{3\times3}$ and the center $x \in \mathbb{R}^3$.
\begin{align}
    G(x)=e^{-1/2(x)^T\Sigma^{-1}(x)} \
\end{align}
The covariance matrix $\Sigma$ of a 3D Gaussian scatter can be analogized to the characterization of an ellipsoidal form. Consequently, it can be further factorized into a rotation matrix denoted as $R\in SO(3)$ and a scale matrix denoted as $S \in \mathbb{R}^3$, permitting independent optimization:
\begin{align}
    \Sigma=RSS^TR^T
\end{align}
To render our 3D Gaussian splats in 2D, we employ the technique of splatting to position the Gaussian splats on the camera planes:
\begin{align}
    \Sigma'=JW\Sigma W^TJ^T
\end{align}
Where J is the Jacobian of the affine approximation of the projective transformation and W is the viewing transformation.
Following this, the pixel color is obtained by alpha-blending N sequentially layered 2D Gaussian splats from front to back:

\begin{align}
    C=\sum_{\substack{i\in N}}c_{i}\alpha_{i}\prod_{j=1}^{i-1}(1-\alpha_{j}) \label{point rendering}
\end{align}

Where $c_{i}$ is the color of each point and $\alpha_{i}$
is given by evaluating a 2D Gaussian with covariance $\Sigma$ multiplied with a learned per-point opacity.

\begin{figure}[htbp]

\includegraphics[width=\linewidth]{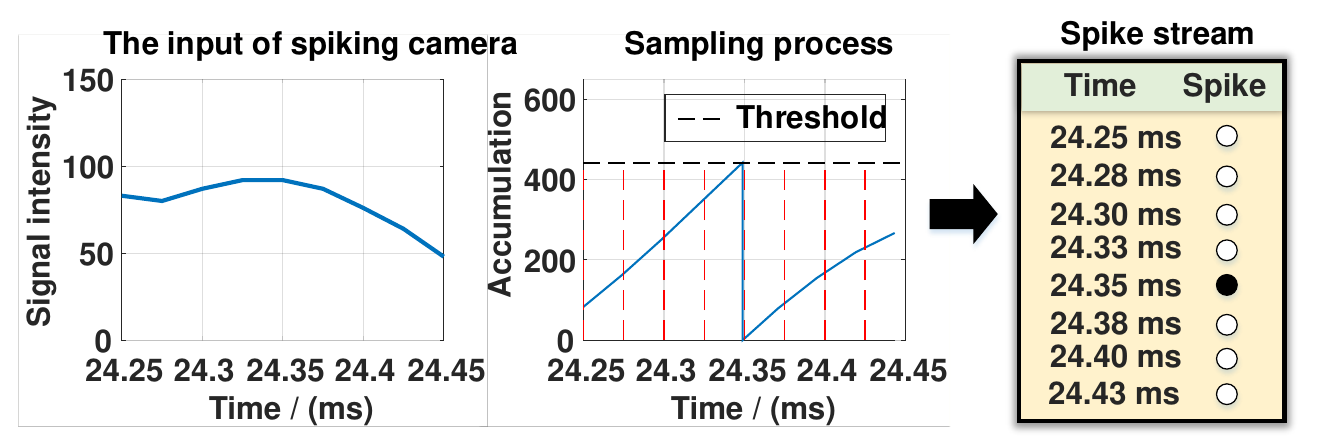}
\centering

\caption{Working principle of each pixel in spike camera. Black (white) circle denotes a (no) spike. }\label{model}

\end{figure}
\subsection{Color Spike Camera Model} 
Each pixel on the color spike camera model converts the light signal into a current signal and accumulates the input current. As shown in Fig.~\ref{model}, for pixel  $\mathbf{x} = (x, y)$, if the accumulation of input current reaches a fixed threshold $\phi$, a spike is fired and then the accumulation can be reset as,
   \begin{align}
    &{A}(\mathbf{x}, t) = {A_\mathbf{x}}(t) \; {\rm{mod}} \; \phi = \int_{0}^{t} {L_{C}}(\mathbf{x}, \tau) d\tau\ {\rm{mod}} \; \phi, 
    \label{eq_spike}
    \end{align}
where ${A}(\mathbf{x}, t)$ is the accumulation at time $t$, ${A_\mathbf{x}}(t)$ is the accumulation without reset before time $t$, ${L_{C}}(\mathbf{x}, \tau)$ is the input current of a certain color at time $\tau$ (proportional to light intensity) and ${C \in \{R,G,B\}}$ denote the color determined by the Bayer-pattern color layout. Note that in a real camera, noise can interfere with the input current ${L_{C}}$ and threshold $\phi$. As shown in Fig.~\ref{model}, each spike is read out at discrete time $nT, n \in \mathbb{N}$ ($T$ is a micro-second level). The output of the color spike camera is a spatial-temporal binary stream $S$ with $H \times W \times N$ size.\textbf{Please refer to appendix for more details}.





\begin{figure}[htbp]
\includegraphics[width=\linewidth]{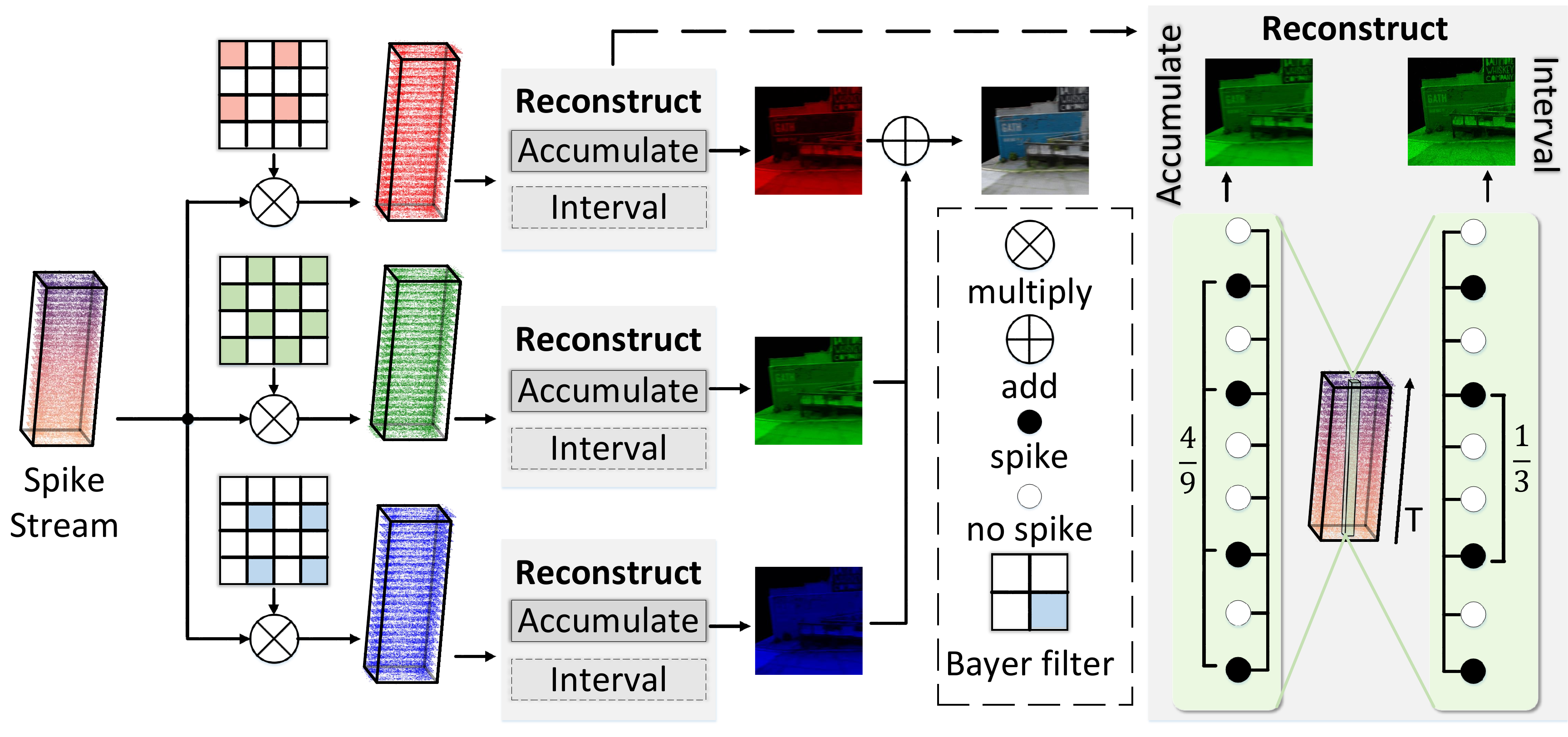}
\centering
\caption{\textbf{Left:} Overview of Bayer-pattern spike stream reconstruction. We employ a Bayer filter to extract the spike
streams of a certain color and calculate accumulating/interval results separately. \textbf{Right: }Reconstruction details. We estimate light intensity by accumulating the number of spikes
over a period of time (equation \ref{eq_TFP1})  or calculating the interval between adjacent spikes (equation \ref{eq_TFI1}).}\label{tfi_tfp}
\vspace{-0.5cm}
\end{figure}
\subsection{Time Accumulation Rasterization}
Due to the instability of the visual texture in a single-frame spike, utilizing the spike stream directly for supervising the training of 3DGS may result in serious multi-view inconsistency and incorrect structure. Therefore, we accumulate long-term temporal information on spike streams to get stable texture. As shown in Fig.~\ref{tfi_tfp}, we mathematically represented this process as follows:
   \begin{align}
        I_{ac}(t_1,t_{N})=\phi/N\sum_{t_i}S(P_{t_i})
    \label{eq_TFP1}
    \end{align}
Where $ I_{ac}(t_1,t_N)$ is the result of spike stream accumulation from $t_1$ to $t_{N}$, $P_{t_i}$ is the camera pose at time $t_i \in  \{t_1,t_2,\dots,t_{N}\}$, $\phi$ is the threshold and $S(P_{t_i})$ is the spike at camera pose $P_{t_i}$.
Additionally, a $2\times2$ Bayer filter, as shown in Fig.~\ref{tfi_tfp}, is employed to extract the spike streams of a certain color and accumulate them separately. The discreet nature of the spike stream in time necessitates the use of summation rather than integration to denote the accumulation process. If the photoelectric conversion process is disregarded, from equation \ref{eq_spike} and equation \ref{eq_TFP1} we have:
\begin{align}
     I_{ac}(t_1,t_{N})=\phi/N\sum_{t_i}\int_{t_{i-1}}^{t_i} {c}(P_{\tau}) d\tau\ {\rm{mod}} \; \phi
    \label{eq_TFP2}
\end{align}
Where ${c}(P_{\tau})$ is the radiance at camera pose $P_{\tau}$. Further we have:
\begin{align}
     I_{ac}(t_1,t_{N})
     \approx
     1/N\int_{t_{1}}^{t_{N}} {c}(P_{\tau}) d\tau\ 
    \label{eq_TFP3}
\end{align}
We have proved in the appendix that the error of the above approximation does not exceed $1/N$. To align the accumulation process of the spike stream, we also improved the rasterization of 3DGS to a time accumulation rasterization by simulating the physical accumulation process of photons: 
\begin{align}
     C_{ac}(t_1,t_{N})
     =
    N^{-1}{\int_{t_{1}}^{t_{N}} \sum_{\substack{i\in N}}c_{i}(P_\tau)\alpha_{i}\prod_{j=1}^{i-1}(1-\alpha_{j}) d\tau\ } 
    \label{eq_TFP4}
\end{align}
Considering computability, equation \ref{eq_TFP4} can be approximated by a discrete summation:
\begin{align}
     C_{ac}(t_1,t_{N})
     =
     N^{-1}{\sum_{\tau=t_{1}}^{t_{N}} \sum_{\substack{i\in N}}c_{i}(P_\tau)\alpha_{i}\prod_{j=1}^{i-1}(1-\alpha_{j})  }
    \label{eq_TFP5}
\end{align}
\begin{figure*}[htbp]

\includegraphics[width=\linewidth]{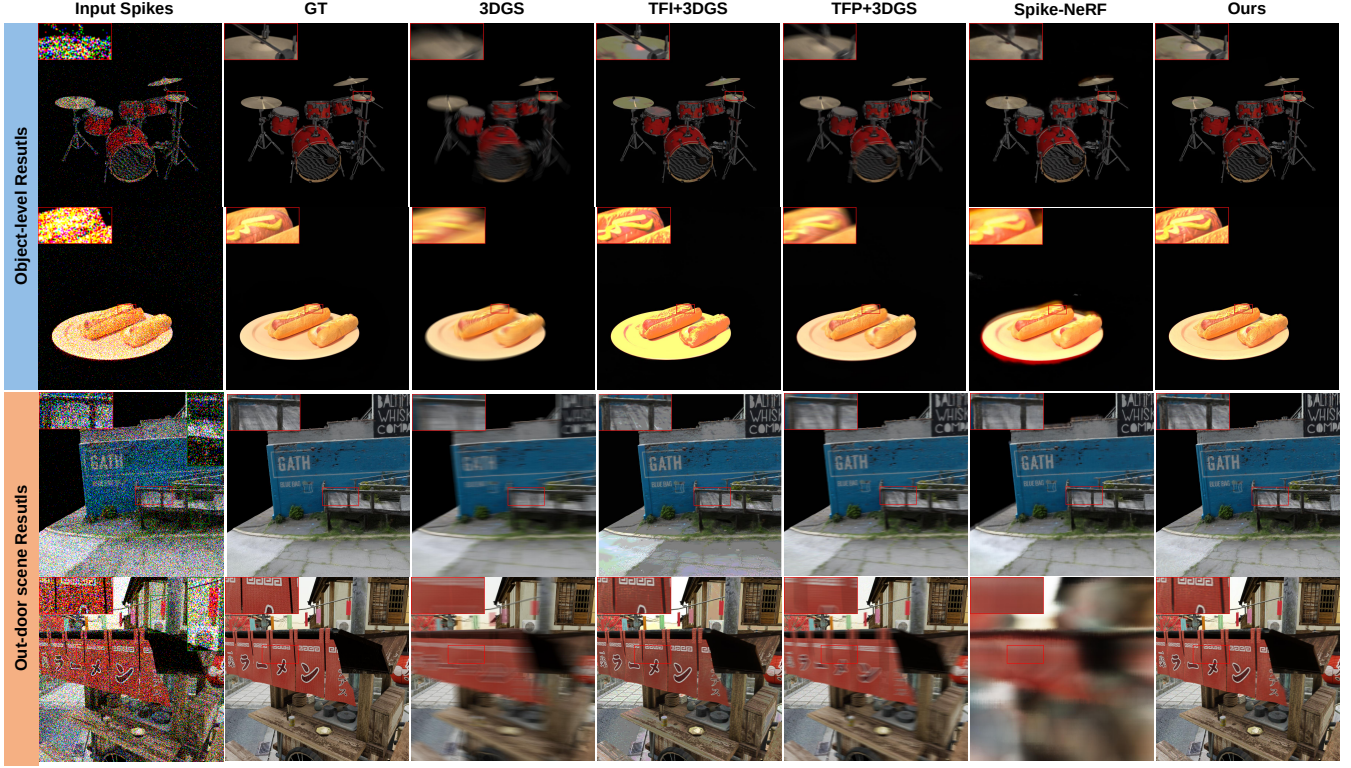}
\centering
\caption{Qualitative comparisons on synthetic datasets. Our SpikeGS achieved sharper, more consistent structures without motion blur like 3DGS or TFP+3DGS  artifacts like Spike-NeRF, and color distortion like TFI+3DGS.}\label{syn_img}
\vspace{-0.1cm}
\end{figure*}
\begin{table*}[htb]
\setlength{\tabcolsep}{0.9mm}
  \centering

    {
    \small
    \begin{tabular}{c|ccc|ccc|ccc}
     \hline
    Datasets  & \multicolumn{3}{c|}{object} & \multicolumn{3}{c|}{outdoor scene}& \multicolumn{3}{c}{average}\\
   Methods$\vert$Metrics &  PSNR $\uparrow$&  SSIM $\uparrow$& LPIPS  $\downarrow$&  PSNR $\uparrow$&  SSIM $\uparrow$& LPIPS  $\downarrow$&  PSNR $\uparrow$&  SSIM $\uparrow$& LPIPS  $\downarrow$\\
     \hline

     3DGS &26.948 &0.880 &0.120&23.379 &0.689 & 0.447&25.163 &0.785 & 0.284 \\
     TFI+3DGS  & \underline{{32.877}} &   0.850 &  \underline{{0.035}} & 26.955 &\underline{{0.823}} &\underline{{0.253}}&\underline{{29.916}} &  0.837 &  \underline{{0.144}}\\
    TFP+3DGS & 30.775& 0.830&0.078& \underline{{27.506}} &0.760 &0.355&29.140 & 0.795 &0.217\\
    S2I+3DGS & 30.421&\underline{{0.946}}&0.060&27.313 &0.762 & 0.284& 28.867 &\underline{{0.854}} & 0.171 \\
    SpikeNeRF &  12.478& 0.828&0.176&11.421 &0.505 &0.460 &11.949 &0.666 & 0.318\\
    Spike-NeRF &   28.243& 0.924&0.081&  18.142&0.621&0.413 &23.192 &0.772 &  0.247\\
    Delur-3DGS &  28.467& 0.918&0.090&23.549 &0.699 &0.412 &26.008 & 0.809 & 0.251\\
    ours & \textbf{{35.860}}& \textbf{{0.960}}&\textbf{{0.030}}& \textbf{{30.915}}& \textbf{{0.856}}&\textbf{{0.210}} &\textbf{{33.387}}& \textbf{{0.908}}&\textbf{{0.120}} \\

    \hline
    \end{tabular}%
}

  \caption{ Quantitative comparisons of novel view synthesis on synthetic datasets. The best and second-best performances are respectively highlighted in bold and underscored with an underline. Our SpikeGS demonstrates the best NVS quality.}
  \label{tab_syn}%

\end{table*}%

\begin{figure*}[ht]

\includegraphics[width=0.95\linewidth]{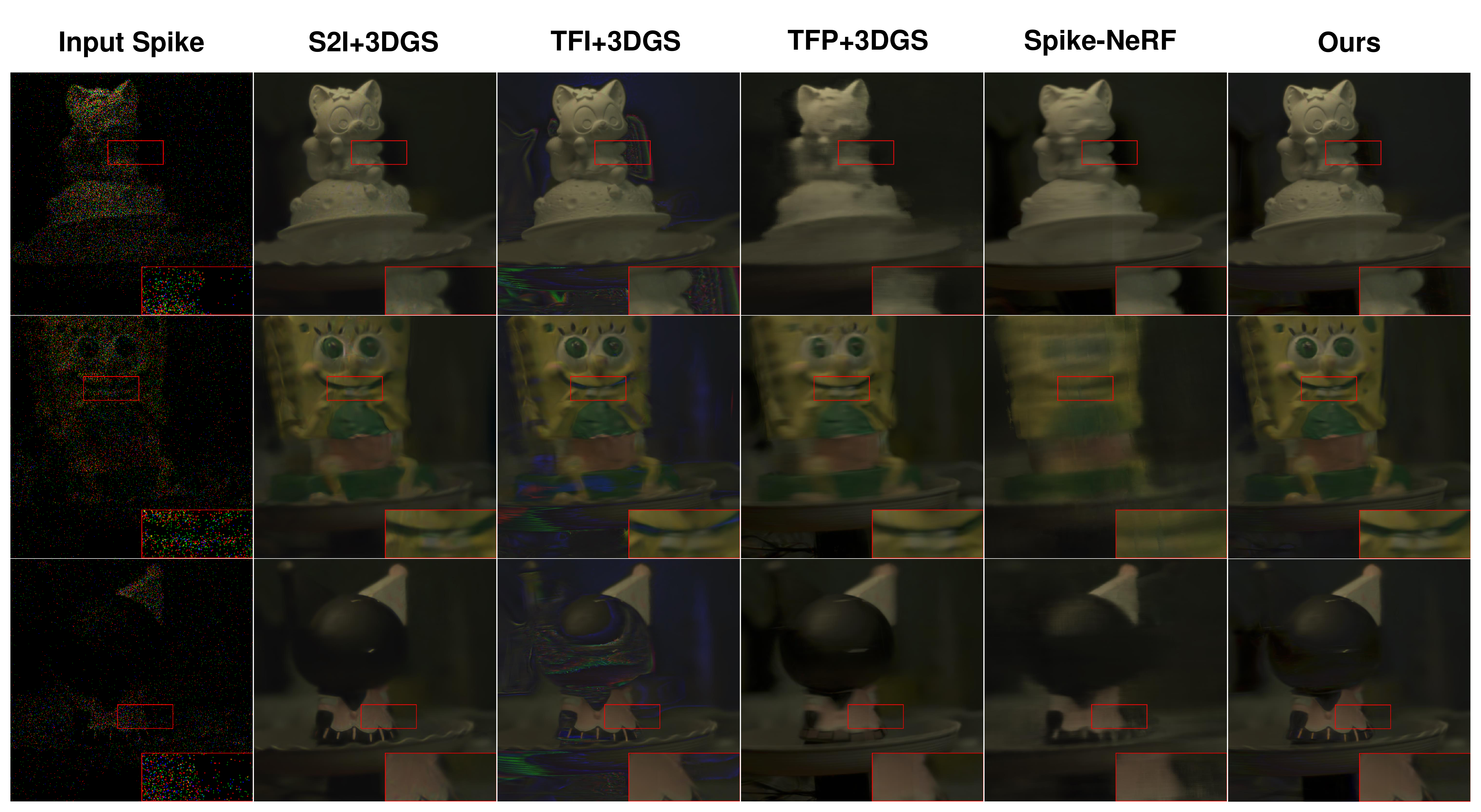}
\centering
\caption{Qualitative comparisons on real-world datasets. Our SpikeGS also significantly outperforms the baseline on real-world datasets. We shows the results from high-speed scene.}\label{real_img}

\end{figure*}
\begin{table*}[h]
  \centering
\small

\setlength{\tabcolsep}{0.85mm}
 {
    \begin{tabular}{c|cc|cc|cc|cc}
    \hline

    Scene Speed  & \multicolumn{2}{c|}{High Speed} & \multicolumn{2}{c|}{Mid Speed}& \multicolumn{2}{c|}{Low Speed}& \multicolumn{2}{c}{Average}   \\
       Methods$\vert$Metrics  &  NIQE $\downarrow$ &  IL-NIQE $\downarrow$ &  NIQE $\downarrow$ &  IL-NIQE $\downarrow$ &  NIQE $\downarrow$ &  IL-NIQE $\downarrow$ &  NIQE $\downarrow$ &  IL-NIQE $\downarrow$\\
       \hline
    TFP+3DGS  & 10.9712 &85.9904&  11.0427& 82.7746&  10.5385&78.5193&  10.8508 &   86.1067\\  
    S2I+3DGS  &   10.9634 & \underline{{78.6108}} &10.9318&  \underline{{73.2949}} &  \underline{{10.4655}}& \underline{{69.8803}}& 10.7869 & \underline{{73.9287}}\\
   Spike-NeRF &   \textbf{{9.6009}} &  85.5263& \underline{{9.4205}}& 83.6021 & 10.6877&78.9751&   \underline{{9.9030}}&  82.7012\\
    Ours  &  \underline{{9.7297}} &  \textbf{{74.8154}}& \textbf{{9.2678}}&   \textbf{{69.8146}}&   \textbf{{8.2760}}&  \textbf{{63.7877}}&  \textbf{{9.0912}} &   \textbf{{69.4726}}\\

    \hline
    \end{tabular}%
}
  \caption{Quantitative comparisons of novel view synthesis on real-world datasets. The best and second-best performances are
respectively highlighted in bold and underscored with an underline. Our SpikeGS demonstrates the best NVS quality.}

  \label{tab_real}%
\end{table*}%
Given the high temporal resolution of the spike stream, the changes in camera pose between consecutive frames are minimal. Consequently, the rendering results between adjacent frames can be assumed to exhibit approximately linear variations. Therefore, the average rendering results of consecutive frames can be approximated by the result from the middle frame. 
In our paper, we use n keyframes (where n $<$ 10) to represent the entire cumulative process. The equation \ref{eq_TFP5} can be further expressed as:
\begin{align}
     C_{ac}(t_1,t_{N})
     =
     n^{-1}{\sum_{i=1}^{n} \sum_{\substack{i\in N}}c_{i}(P_i)\alpha_{i}\prod_{j=1}^{i-1}(1-\alpha_{j})  } 
    \label{eq_TFP6}
\end{align}
Where $P_i$ is the camera pose at the keyframe i. The above process is fully differentiable. The derivation of differentiable rendering is also provided in the appendix. Finally, we employ an accumulation loss to minimize their photometric error, expressed as:
\begin{align}
    \mathcal L_{ac}=(1-\lambda_1)\Vert C_{ac}-I_{ac} \Vert_2+\lambda_1 SSIM(C_{ac},I_{ac})
\label{loss_acc}
\end{align}


\subsection{Mutually Constrained Training and Loss Function}
While using time accumulation rasterization can help recover texture details and geometric information effectively, it tends to hinder the rapid convergence of Gaussian splats during the early stages of training, significantly impacting training efficiency. Meanwhile, the accuracy of point cloud initialization and the geometric precision during the early stages of Gaussian splats training have a significant impact on the final reconstruction quality of 3DGS. We found that whether using a single frame spike or accumulated spike streams, the results of point cloud initialization and the initial training were unsatisfactory. In our paper, we use spike intervals, which have higher sensitivity and flexibility to the time domain, to initialize point clouds and use them for initial training, as shown in Fig.~\ref{pipeline}. In order to obtain more accurate color and texture information from the spike interval \cite{zhu2019retina}, we added additional processing to the spike interval, just like Fig.~\ref{tfi_tfp}:
   \begin{align}
        I_{in}=\frac{\phi}{t_1-t_2}
    \label{eq_TFI1}
    \end{align}
Where $\phi$ represents the threshold. $t_1$ and $t_2$ represent the times corresponding to two adjacent trigger spikes ($S(t_1)=S(t_2)=1$). We used a 2 × 2 Bayer filter the same as accumulation. Using the spike interval allows the recovery of geometric details using minimal numbers of spikes (typically t1-t2 $<$10). However, spike intervals have significant time-domain aliasing, which manifests as color deviations and is highly susceptible to noise interference. Therefore, we gradually reduce its impact during later stages of training, and ultimately only use it to adjust the geometric parameters of the Gaussian, such as position and opacity. Rendering results corresponding to the spike intervals can be represented as:
\begin{align}
     C_{in}
     =
      \sum_{\substack{i\in N}}c_{i}(P_{(t_1+t_2)/2})\alpha_{i}\prod_{j=1}^{i-1}(1-\alpha_{j})
    \label{eq_TFI2}
\end{align}
We also employ an interval loss which can be expressed as :
\begin{align}
    \mathcal L_{in}=(1-\lambda_1)\Vert C_{in}-I_{in} \Vert_2+\lambda_1 SSIM(C_{in},I_{in})
\label{loss_interval}
\end{align}
Our final loss function can be represented as:
\begin{align}
    \mathcal L_{final}=\lambda_{accu}\mathcal L_{in}+\lambda_{in}\mathcal L_{in}
\label{final_loss}
\end{align}
Where $\lambda_{accu}$ and $\lambda_{in}$ are hyperparameters. $\lambda_{accu}$ starts at 0 in the early stages of training and gradually increases with the number of epochs.

\vspace{-0.4cm}
\section{Experiment}

\begin{table*}[htbp]
  \centering

\setlength{\tabcolsep}{0.85mm}
\small

 {
    \begin{tabular}{c|ccc|ccc|ccc}
    \hline

    Scene  & \multicolumn{3}{c|}{High Speed} & \multicolumn{3}{c|}{Low Speed}& \multicolumn{3}{c}{Average}   \\
      Method$|$Metrics  &  PSNR $\uparrow$ &  SSIM $\uparrow$& LPIPS  $\downarrow$&  PSNR $\uparrow$ &  SSIM $\uparrow$& LPIPS  $\downarrow$&  PSNR $\uparrow$ &  SSIM $\uparrow$& LPIPS  $\downarrow$\\
      \hline
    3DGS &   25.225 &  0.850& 0.139&  26.948 &  0.880& 0.120&  26.08 & 0.865&0.130\\
    Deblur-3DGS&   26.422 & \underline{{0.897}}&  0.110& 28.467&   0.918& 0.090&  27.44 &   0.907& 0.100\\ 
    Ours w/o $\mathcal L_{in}$ &32.50 &  0.868&  0.063&  \underline{{35.75}}& \underline{{0.948}}& \underline{{0.031}}& 34.12 &  0.908& 0.047\\
    Ours  w/o $\mathcal L_{accu}$ &   29.95 &  0.859& 0.088& 32.42 & 0.943& 0.063&  31.18 &  0.901& 0.075\\
    Ours-3 views  &  31.90 & 0.862&0.061&  34.19&   0.940& 0.043& 33.04 &  0.901& 0.052\\
    Ours-5 views  &  \underline{{32.94}} &  0.870&  \underline{{0.052}}& {35.34} & 0.947& 0.035& \underline{{34.14}} &  \underline{{0.908}}& \underline{{0.043}}\\
    Ours-9 views  &  \textbf{{33.45}} & \textbf{{0.921}}& \textbf{{0.0467}}&  \textbf{{35.86}} &  \textbf{{0.960}}& \textbf{{0.030}}&  \textbf{{34.65}}&  \textbf{{0.941}}& \textbf{{0.038}}\\

    \hline
    \end{tabular}%
}

  \caption{Quantitative ablation on  $\mathcal L_{in}$, $\mathcal L_{accu}$, and numbers of camera poses on scenes with different speeds. The best and second-best performances are
respectively highlighted in bold and underscored with an underline. All of them have a great influence on NVS quality.}

  \label{tab_ab}%
\end{table*}

\begin{table}[htb]
  \centering

    \setlength{\tabcolsep}{0.85mm}
\small
    {
    \begin{tabular}{c|ccc}
     \hline
    
   Methods$\vert$Metrics &  PSNR $\uparrow$&  SSIM $\uparrow$& LPIPS  $\downarrow$\\
     \hline

     3DGS&18.757 &0.885 &0.137 \\
     TFI+3DGS   &\underline{{22.72}} &0.327 &0.096\\
    TFP+3DGS & 22.69 &0.729 &0.104\\
    S2I+3DGS &21.41 &\underline{{0.912}} & \underline{{0.094}} \\
    SpikeNeRF &10.42 &0.103 &0.260 \\
    Spike-NeRF&11.72 &0.280 &0.220 \\
    Delur-3DGS&20.98 &0.843 &0.115 \\
    ours & \textbf{{27.360}}& \textbf{{0.913}}&\textbf{{0.059}} \\

    \hline
    \end{tabular}%
}
  \caption{ Quantitative comparisons of Depth Estimation on synthetic datasets. The best and second-best performances are
respectively highlighted in bold and underscored with an underline. Our SpikeGS demonstrates the best quality.}

  \label{tab_depth}%
\end{table}%

\label{sec:experiment}
\subsection{Implementation Details}
Our code is based on 3DGS \cite{3dgs} and we train the models for $30000$ iterations on one NVIDIA A800 GPU with the same optimizer and hyper-parameters as 3DGS. The color spike camera shown in Fig.~\ref{setup} is \textbf{Spike M1K40-H4-Gen3} with $1000\times1000$ resolution and the motor shown in Fig.~\ref{setup} is \textbf{DMR4315}. 

\subsection{Experimental Settings}

We first compare our method with 3D-GS \cite{3dgs} and deblur-GS \cite{chen2024deblur} directly trained on the blurry images. Then, we compare our method against several 3D scene reconstruction methods for spike cameras, such as SpikeNeRF \cite{zhu2024spikenerf} and Spike-NeRF \cite{guo2024spike}. Finally, given the absence of existing spike-based 3DGS methods, we compared our method with three spike-based image reconstruction methods with 3DGS: TFI \cite{zhu2019retina}+3DGS, TFP \cite{zhu2019retina}+3DGS and Spk2imgNet \cite{zhao2021spk2imgnet} +3DGS.
We conduct experiments using proposed SpikeGS datasets to evaluate our method's performance on both synthetic and real-world scenes. For synthetic datasets, we evaluate the
synthesized novel view in terms of PSNR, SSIM, and LPIPS. For real-world datasets, we evaluate the synthesized novel view using a set of No-Reference Image Quality Assessment (NR-IQA) metrics since we don't have ground truth. The adopted metrics include NIQE \cite{mittal2012making} and IL-NIQE \cite{zhang2015feature}. 

\subsection{Comparisons on Synthetic Datasets}
\begin{figure}[ht]
\includegraphics[width=\linewidth]{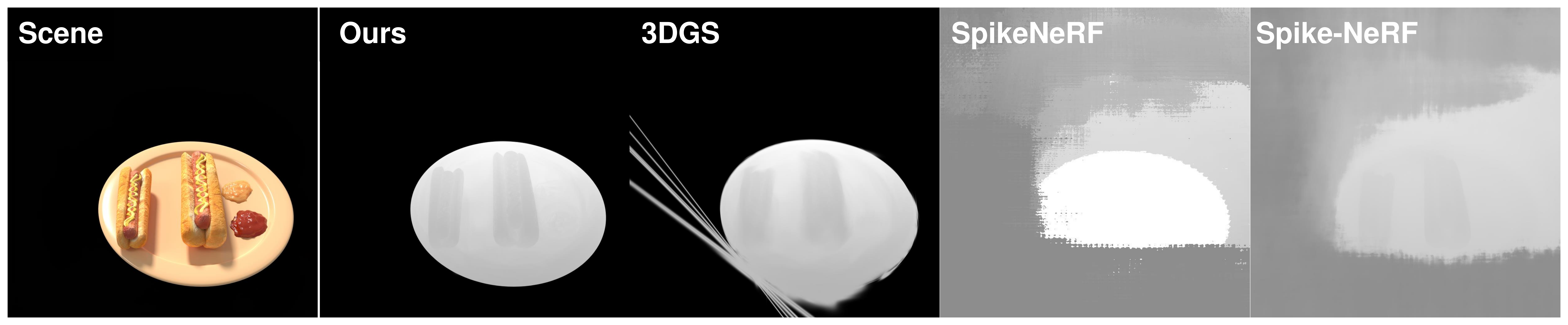}
\centering
\caption{Qualitative comparisons on depth estimation. Our SpikeGS achieves significantly better structural details. }\label{depth_img}
\end{figure}

We compare our method with the aforementioned baselines on synthetic datasets. The results in Tab.~\ref{tab_syn} indicate that our SpikeGS achieved state-of-the-art novel view synthesis quality compared to other 3D reconstruction methods based on spike cameras and the deblur method based on traditional RGB camera. On average, our method achieves a +3.47dB higher PSNR, a 6.32\% higher SSIM and a 14\% lower LPIPS compared to the best performance among all baselines. We also perform qualitative experiments on synthetic data. Fig.~\ref{syn_img} demonstrated that our SpikeGS achieved sharper, more consistent structures. Meanwhile, we validated the accuracy of the geometry by comparing the precision of depth estimation. Tab.~\ref{tab_depth} shows that the geometric results reconstructed by our SpikeGS method significantly outperforms baselines. Visual results are shown in Fig \ref{depth_img}.
\subsection{Comparisons on Real-world Datasets}
Due to the lack of data from other modalities, we did not compare with non-spike methods on real-world data. 
According to Table \ref{tab_real}, our approach demonstrated superior performance compared to other spike-based methods across three different scene speeds: high, medium, and low. This outcome validates the applicability of our method across a broad range of speeds.
Fig.~\ref{real_img} demonstrates that our SpikeGS achieves more visually appealing results without introducing blurring, high-frequency noise, and erroneous backgrounds.

\subsection{Ablation}
\textbf{Numbers of Camera Poses.} We conduct ablations to investigate the effect of the number of camera poses (key frames) optimized in the accumulation. We select two speeds from our synthetic dataset for evaluation. As depicted in Tab.~\ref{tab_ab}, the number of camera poses had a significant impact on the quality of reconstruction results, and the impact is more pronounced at high speeds. Here, we
also provide a comparison with Deblur-3DGS (10 views) \cite{chen2024deblur} and 3DGS \cite{3dgs}, which demonstrates that our approach demonstrates spectacular superiority over methods based on traditional RGB camera.
\\\textbf{Accumulation Loss and Interval Loss.} We also conduct novel view synthesis
experiments to investigate the effect of interval and accumulation on the same dataset. Tab.~\ref{tab_ab} shows that both interval and accumulation loss significantly enhance novel view synthesis quality. Moreover, the impact of accumulation loss remains consistent across both low-speed and high-speed scenarios while the influence of interval loss is notably more pronounced in high-speed scenarios. 

\section{Conclusion}
\label{sec:conclusion}
This paper introduces Spike Gaussian Splatting (SpikeGS), the first framework that integrates the Bayer-pattern 
spike streams into the 3DGS pipeline to reconstruct a 3D scene via a fast-moving spike camera. With accumulation rasterization, interval supervision, and a specially designed pipeline, SpikeGS reconstructs detailed geometry and texture from spike streams which are unstable and lack details, addressing the challenges associated with real-world Bayer-pattern spike streams. Evaluation of both synthetic and real-world datasets with multi-scenes and multi-speed demonstrate that our SpikeGS achieved state-of-the-art novel view synthesis quality compared to other spike-based methods. 

\section{Acknowledgments}
This work was supported by National Scienceand Technology Major Project (2022ZD0116305)

\bibliography{aaai25}

\begin{thebibliography}{34}
\providecommand{\natexlab}[1]{#1}

\bibitem[{Barron et~al.(2021)Barron, Mildenhall, Tancik, Hedman, Martin-Brualla, and Srinivasan}]{barron2021mipnerf}
Barron, J.~T.; Mildenhall, B.; Tancik, M.; Hedman, P.; Martin-Brualla, R.; and Srinivasan, P.~P. 2021.
\newblock Mip-nerf: A multiscale representation for anti-aliasing neural radiance fields.
\newblock In \emph{Proceedings of the IEEE/CVF International Conference on Computer Vision}, 5855--5864.

\bibitem[{Barron et~al.(2022)Barron, Mildenhall, Verbin, Srinivasan, and Hedman}]{barron2022mipnerf360}
Barron, J.~T.; Mildenhall, B.; Verbin, D.; Srinivasan, P.~P.; and Hedman, P. 2022.
\newblock Mip-nerf 360: Unbounded anti-aliased neural radiance fields.
\newblock In \emph{Proceedings of the IEEE/CVF Conference on Computer Vision and Pattern Recognition}, 5470--5479.

\bibitem[{Chen et~al.(2022)Chen, Xu, Geiger, Yu, and Su}]{chen2022tensorf}
Chen, A.; Xu, Z.; Geiger, A.; Yu, J.; and Su, H. 2022.
\newblock Tensorf: Tensorial radiance fields.
\newblock In \emph{European Conference on Computer Vision}, 333--350. Springer.

\bibitem[{Chen and Liu(2024)}]{chen2024deblur}
Chen, W.; and Liu, L. 2024.
\newblock Deblur-GS: 3D Gaussian Splatting from Camera Motion Blurred Images.
\newblock \emph{Proceedings of the ACM on Computer Graphics and Interactive Techniques}, 7(1): 1--15.

\bibitem[{Chen et~al.(2023)Chen, Chen, Zhang, Wang, Yang, Wang, Cai, Yang, Liu, and Lin}]{chen2023gaussianeditor}
Chen, Y.; Chen, Z.; Zhang, C.; Wang, F.; Yang, X.; Wang, Y.; Cai, Z.; Yang, L.; Liu, H.; and Lin, G. 2023.
\newblock Gaussianeditor: Swift and controllable 3d editing with gaussian splatting.
\newblock \emph{arXiv preprint arXiv:2311.14521}.

\bibitem[{Dong, Huang, and Tian(2021)}]{dong2021spike}
Dong, S.; Huang, T.; and Tian, Y. 2021.
\newblock Spike camera and its coding methods.
\newblock \emph{arXiv preprint arXiv:2104.04669}.

\bibitem[{Fan et~al.(2024)Fan, Cong, Wen, Wang, Zhang, Ding, Xu, Ivanovic, Pavone, Pavlakos et~al.}]{fan2024instantsplat}
Fan, Z.; Cong, W.; Wen, K.; Wang, K.; Zhang, J.; Ding, X.; Xu, D.; Ivanovic, B.; Pavone, M.; Pavlakos, G.; et~al. 2024.
\newblock Instantsplat: Unbounded sparse-view pose-free gaussian splatting in 40 seconds.
\newblock \emph{arXiv preprint arXiv:2403.20309}.

\bibitem[{Fridovich-Keil et~al.(2022)Fridovich-Keil, Yu, Tancik, Chen, Recht, and Kanazawa}]{fridovich2022plenoxels}
Fridovich-Keil, S.; Yu, A.; Tancik, M.; Chen, Q.; Recht, B.; and Kanazawa, A. 2022.
\newblock Plenoxels: Radiance fields without neural networks.
\newblock In \emph{Proceedings of the IEEE/CVF Conference on Computer Vision and Pattern Recognition}, 5501--5510.

\bibitem[{Guo et~al.(2024{\natexlab{a}})Guo, Bai, Hu, Guo, Liu, Cai, Huang, and Ma}]{guo2024prtgs}
Guo, Y.; Bai, Y.; Hu, L.; Guo, Z.; Liu, M.; Cai, Y.; Huang, T.; and Ma, L. 2024{\natexlab{a}}.
\newblock PRTGS: Precomputed Radiance Transfer of Gaussian Splats for Real-Time High-Quality Relighting.
\newblock \emph{arXiv preprint arXiv:2408.03538}.

\bibitem[{Guo et~al.(2024{\natexlab{b}})Guo, Bai, Hu, Liu, Guo, Ma, and Huang}]{guo2024spike}
Guo, Y.; Bai, Y.; Hu, L.; Liu, M.; Guo, Z.; Ma, L.; and Huang, T. 2024{\natexlab{b}}.
\newblock Spike-NeRF: Neural Radiance Field Based On Spike Camera.
\newblock \emph{arXiv preprint arXiv:2403.16410}.

\bibitem[{Hu et~al.(2024)Hu, Ma, Guo, and Huang}]{hu2024scsimrealisticspikecameras}
Hu, L.; Ma, L.; Guo, Y.; and Huang, T. 2024.
\newblock SCSim: A Realistic Spike Cameras Simulator.
\newblock arXiv:2405.16790.

\bibitem[{Huang et~al.(2023)Huang, Zheng, Yu, Chen, Li, Xiong, Ma, Zhao, Dong, Zhu et~al.}]{huang20231000}
Huang, T.; Zheng, Y.; Yu, Z.; Chen, R.; Li, Y.; Xiong, R.; Ma, L.; Zhao, J.; Dong, S.; Zhu, L.; et~al. 2023.
\newblock 1000$\times$ faster camera and machine vision with ordinary devices.
\newblock \emph{Engineering}, 25: 110--119.

\bibitem[{Hwang, Kim, and Kim(2023)}]{hwang2023ev}
Hwang, I.; Kim, J.; and Kim, Y.~M. 2023.
\newblock Ev-NeRF: Event based neural radiance field.
\newblock In \emph{Proceedings of the IEEE/CVF Winter Conference on Applications of Computer Vision}, 837--847.

\bibitem[{Kerbl et~al.(2023)Kerbl, Kopanas, Leimk{\"u}hler, and Drettakis}]{3dgs}
Kerbl, B.; Kopanas, G.; Leimk{\"u}hler, T.; and Drettakis, G. 2023.
\newblock 3d gaussian splatting for real-time radiance field rendering.
\newblock \emph{ACM Transactions on Graphics}, 42(4): 1--14.

\bibitem[{Liang et~al.(2023)Liang, Zhang, Feng, Shan, and Jia}]{liang2023gsir}
Liang, Z.; Zhang, Q.; Feng, Y.; Shan, Y.; and Jia, K. 2023.
\newblock Gs-ir: 3d gaussian splatting for inverse rendering.
\newblock \emph{arXiv preprint arXiv:2311.16473}.

\bibitem[{Low and Lee(2023)}]{low2023robust}
Low, W.~F.; and Lee, G.~H. 2023.
\newblock Robust e-NeRF: NeRF from Sparse \& Noisy Events under Non-Uniform Motion.
\newblock In \emph{Proceedings of the IEEE/CVF International Conference on Computer Vision}, 18335--18346.

\bibitem[{Ma et~al.(2022)Ma, Li, Liao, Zhang, Wang, Wang, and Sander}]{ma2022deblur}
Ma, L.; Li, X.; Liao, J.; Zhang, Q.; Wang, X.; Wang, J.; and Sander, P.~V. 2022.
\newblock Deblur-nerf: Neural radiance fields from blurry images.
\newblock In \emph{Proceedings of the IEEE/CVF Conference on Computer Vision and Pattern Recognition}, 12861--12870.

\bibitem[{Ma et~al.(2023)Ma, Paudel, Chhatkuli, and Van~Gool}]{ma2023deformable}
Ma, Q.; Paudel, D.~P.; Chhatkuli, A.; and Van~Gool, L. 2023.
\newblock Deformable Neural Radiance Fields using RGB and Event Cameras.
\newblock In \emph{Proceedings of the IEEE/CVF International Conference on Computer Vision}, 3590--3600.

\bibitem[{Mildenhall et~al.(2021)Mildenhall, Srinivasan, Tancik, Barron, Ramamoorthi, and Ng}]{nerf}
Mildenhall, B.; Srinivasan, P.~P.; Tancik, M.; Barron, J.~T.; Ramamoorthi, R.; and Ng, R. 2021.
\newblock Nerf: Representing scenes as neural radiance fields for view synthesis.
\newblock \emph{Communications of the ACM}, 65(1): 99--106.

\bibitem[{Mittal, Soundararajan, and Bovik(2012)}]{mittal2012making}
Mittal, A.; Soundararajan, R.; and Bovik, A.~C. 2012.
\newblock Making a “completely blind” image quality analyzer.
\newblock \emph{IEEE Signal processing letters}, 20(3): 209--212.

\bibitem[{M{\"u}ller et~al.(2022)M{\"u}ller, Evans, Schied, and Keller}]{muller2022instant}
M{\"u}ller, T.; Evans, A.; Schied, C.; and Keller, A. 2022.
\newblock Instant neural graphics primitives with a multiresolution hash encoding.
\newblock \emph{ACM transactions on graphics (TOG)}, 41(4): 1--15.

\bibitem[{Qi et~al.(2023)Qi, Zhu, Zhang, and Li}]{qi2023e2nerf}
Qi, Y.; Zhu, L.; Zhang, Y.; and Li, J. 2023.
\newblock E2nerf: Event enhanced neural radiance fields from blurry images.
\newblock In \emph{Proceedings of the IEEE/CVF International Conference on Computer Vision}, 13254--13264.

\bibitem[{Rudnev et~al.(2023)Rudnev, Elgharib, Theobalt, and Golyanik}]{rudnev2023eventnerf}
Rudnev, V.; Elgharib, M.; Theobalt, C.; and Golyanik, V. 2023.
\newblock EventNeRF: Neural radiance fields from a single colour event camera.
\newblock In \emph{Proceedings of the IEEE/CVF Conference on Computer Vision and Pattern Recognition}, 4992--5002.

\bibitem[{Wang et~al.(2021)Wang, Wu, Xie, Chen, and Prisacariu}]{wang2021nerf--}
Wang, Z.; Wu, S.; Xie, W.; Chen, M.; and Prisacariu, V.~A. 2021.
\newblock NeRF--: Neural radiance fields without known camera parameters.
\newblock \emph{arXiv preprint arXiv:2102.07064}.

\bibitem[{Wu et~al.(2023)Wu, Yi, Fang, Xie, Zhang, Wei, Liu, Tian, and Wang}]{wu20234d}
Wu, G.; Yi, T.; Fang, J.; Xie, L.; Zhang, X.; Wei, W.; Liu, W.; Tian, Q.; and Wang, X. 2023.
\newblock 4d gaussian splatting for real-time dynamic scene rendering.
\newblock \emph{arXiv preprint arXiv:2310.08528}.

\bibitem[{Xiong et~al.(2024)Xiong, Wu, He, Fermuller, Aloimonos, Huang, and Metzler}]{xiong2024event3dgs}
Xiong, T.; Wu, J.; He, B.; Fermuller, C.; Aloimonos, Y.; Huang, H.; and Metzler, C.~A. 2024.
\newblock Event3DGS: Event-based 3D Gaussian Splatting for Fast Egomotion.
\newblock \emph{arXiv preprint arXiv:2406.02972}.

\bibitem[{Yang et~al.(2023)Yang, Yang, Pan, Zhu, and Zhang}]{yang2023real}
Yang, Z.; Yang, H.; Pan, Z.; Zhu, X.; and Zhang, L. 2023.
\newblock Real-time photorealistic dynamic scene representation and rendering with 4d gaussian splatting.
\newblock \emph{arXiv preprint arXiv:2310.10642}.

\bibitem[{Yi et~al.(2023)Yi, Fang, Wu, Xie, Zhang, Liu, Tian, and Wang}]{yi2023gaussiandreamer}
Yi, T.; Fang, J.; Wu, G.; Xie, L.; Zhang, X.; Liu, W.; Tian, Q.; and Wang, X. 2023.
\newblock Gaussiandreamer: Fast generation from text to 3d gaussian splatting with point cloud priors.
\newblock \emph{arXiv preprint arXiv:2310.08529}.

\bibitem[{Zhang et~al.(2020)Zhang, Riegler, Snavely, and Koltun}]{zhang2020nerf++}
Zhang, K.; Riegler, G.; Snavely, N.; and Koltun, V. 2020.
\newblock Nerf++: Analyzing and improving neural radiance fields.
\newblock \emph{arXiv preprint arXiv:2010.07492}.

\bibitem[{Zhang, Zhang, and Bovik(2015)}]{zhang2015feature}
Zhang, L.; Zhang, L.; and Bovik, A.~C. 2015.
\newblock A feature-enriched completely blind image quality evaluator.
\newblock \emph{IEEE Transactions on Image Processing}, 24(8): 2579--2591.

\bibitem[{Zhao et~al.(2021)Zhao, Xiong, Liu, Zhang, and Huang}]{zhao2021spk2imgnet}
Zhao, J.; Xiong, R.; Liu, H.; Zhang, J.; and Huang, T. 2021.
\newblock Spk2imgnet: Learning to reconstruct dynamic scene from continuous spike stream.
\newblock In \emph{Proceedings of the IEEE/CVF Conference on Computer Vision and Pattern Recognition}, 11996--12005.

\bibitem[{Zhao, Wang, and Liu(2024)}]{zhao2024bad}
Zhao, L.; Wang, P.; and Liu, P. 2024.
\newblock Bad-gaussians: Bundle adjusted deblur gaussian splatting.
\newblock \emph{arXiv preprint arXiv:2403.11831}.

\bibitem[{Zhu et~al.(2019)Zhu, Dong, Huang, and Tian}]{zhu2019retina}
Zhu, L.; Dong, S.; Huang, T.; and Tian, Y. 2019.
\newblock A retina-inspired sampling method for visual texture reconstruction.
\newblock In \emph{2019 IEEE International Conference on Multimedia and Expo (ICME)}, 1432--1437. IEEE.

\bibitem[{Zhu et~al.(2024)Zhu, Jia, Zhao, Qi, Wang, and Huang}]{zhu2024spikenerf}
Zhu, L.; Jia, K.; Zhao, Y.; Qi, Y.; Wang, L.; and Huang, H. 2024.
\newblock SpikeNeRF: Learning Neural Radiance Fields from Continuous Spike Stream.
\newblock In \emph{Proceedings of the IEEE/CVF Conference on Computer Vision and Pattern Recognition}, 6285--6295.

\end{thebibliography}

\end{document}